\pdfoutput=1
\documentclass{article}

\usepackage[preprint]{corl_2026}
\makeatletter
\renewcommand{\@noticestring}{CoRL 2026 Workshop ``Pretrain to Adapt: What Makes a Pretrained Policy Adaptable?'', \@conferencelocation.}
\makeatother
\hypersetup{
  pdftitle={VLA Adaptation with Real and Simulated Data: A Case Study on Robotic Chess},
  pdfauthor={Matej Kripner, Milan Straka},
  pdfsubject={CoRL 2026 Workshop Pretrain to Adapt},
}

\usepackage{booktabs}
\usepackage{graphicx}
\usepackage{float}
\usepackage{amsmath}
\usepackage{xcolor}
\graphicspath{{figures/}}

\newcommand{\pizerofive}{$\pi_{0.5}$}

\title{VLA Adaptation with Real and Simulated Data:\\A Case Study on Robotic Chess}

\author{
  Mat\v{e}j Kripner \qquad Milan Straka \\
  Charles University, Faculty of Mathematics and Physics \\
  Prague, Czech Republic \\
  \texttt{kripner@ufal.mff.cuni.cz}
}

\begin{document}
\maketitle

\begin{abstract}
Vision-language-action (VLA) models are trained as general-purpose foundation models for robot learning, and require fine-tuning for each downstream task and robot.
This raises practical questions such as how much data and training updates are needed, how checkpoints should be selected, and whether simulation can be leveraged to alleviate the cost of real data collection.
We study these questions on a task that requires both precise manipulation and spatial understanding of the scene.
In this task, an SO-101 arm moves a chess king between two squares that are given only by their names, such as e4 and a5.
We fine-tune \pizerofive{} with various dataset sizes and update budgets, using teleoperated demonstrations as well as simulated demonstrations from a randomized scripted expert that imitates the variability of human demonstrations.
In simulation, the success rate scales with both data and updates.
Importantly, we find that a short learning-rate cool-down adds up to 31 points and is necessary for comparing checkpoints, and that validation loss and action error are unreliable for model selection.
On the real arm, we fine-tune on up to 256 teleoperated demonstrations and evaluate ten policies in 25 trials each.
Partial results indicate that pretraining on the simulated demonstrations helps substantially, most clearly when real data is scarce.
We also present a negative result, in which training on the policy's own rollouts with filtered behavior cloning, RECAP-style advantage conditioning, corrections from an automatic coach, or residual reinforcement learning did not yield a measurable improvement over plain behavior cloning in our setting.
We release the simulator, the expert and the real demonstrations.
\end{abstract}

\section{Introduction}
\label{sec:intro}

A common way to teach a robot a new manipulation task is to fine-tune a pretrained vision-language-action (VLA) model~\citep{pi0,pi05,smolvla} on human-collected demonstrations of that task.
This raises practical questions, such as how many demonstrations to collect, how long to train, which checkpoint to deploy, and whether pretraining in simulation is worthwhile.
Each of these choices carries a substantial cost in data, compute or robot time.

A central challenge is that closed-loop task success, the quantity of interest, is expensive to measure.
On a real robot, each trial takes minutes and requires a person to reset the scene.
It is therefore tempting to rely on offline proxies such as validation loss or action error, or on a few dozen trials per setting.
Our study complements the fine-tuning experiments reported with the VLA models themselves~\citep{pi0,pi05,smolvla} and data-scaling studies of imitation learning~\citep{lin2024datascaling} with a detailed look at a single task, where a simulator lets us evaluate every model on many trials.

\begin{figure}[t]
  \centering
  \includegraphics{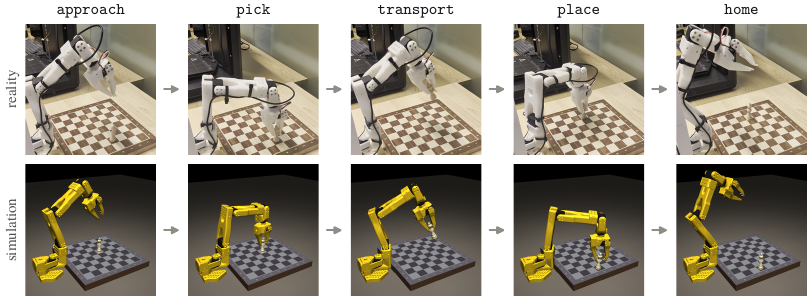}
  \caption{The five stages of a move on the real arm (top) and in simulation (bottom), labeled with the stage tokens the policy predicts.}
  \label{fig:stages}
\end{figure}

In this paper, we study the task of moving a chess king on a board with a low-cost robot arm.
Given an instruction such as \emph{``Move the king from e4 to a5,''} the policy must locate the named squares on the board it sees and then grasp, carry and place the king (Fig.~\ref{fig:stages}).
The task therefore requires both precise manipulation and spatial understanding of the scene.
We fine-tune \pizerofive{}~\citep{pi05} on teleoperated demonstrations and on simulated demonstrations from a randomized scripted expert.
The simulator provides demonstrations in any quantity and lets us evaluate every model on the same 1,000 graded trials.
On the real arm, we fine-tune on up to 256 teleoperated demonstrations, starting either from the base \pizerofive{} model or from a policy pretrained on the simulated demonstrations.
We also test whether the policy improves when it is trained on its own rollouts.
Beyond king moves, we also built a system on the same arm that plays full games with all pieces, but only as a demonstration, without evaluation or ablations.

In summary, our contributions are:
\begin{itemize}
  \item In a simulated version of the task, we present quantitative experiments indicating that the success rate scales with both data and updates, that checkpoints must be compared after a learning-rate cool-down or stochastic weight averaging (SWA)~\citep{swa}, and that validation loss and action error are unreliable for model selection (Section~\ref{sec:bc}).
  \item On the real arm, partial results indicate that pretraining on simulated demonstrations helps when real data is scarce (Section~\ref{sec:real}).
  \item As a negative result, training on the policy's own rollouts with filtered behavior cloning, RECAP-style advantage conditioning~\citep{pistar06}, corrections from an automatic coach or residual reinforcement learning did not yield a measurable improvement over behavior cloning in our setting (Section~\ref{sec:experience}).
  \item We release the simulator with its randomized scripted expert and the teleoperated demonstrations, and contribute the recordings of the real evaluation trials to the Oopsie Data collection of real-robot policy rollouts\footnote{\url{https://oopsie-data.com}} (Section~\ref{sec:sim}, Appendix~\ref{app:data}).
\end{itemize}

\section{Setup}
\label{sec:setup}

\begin{figure}[t]
  \centering
  \includegraphics{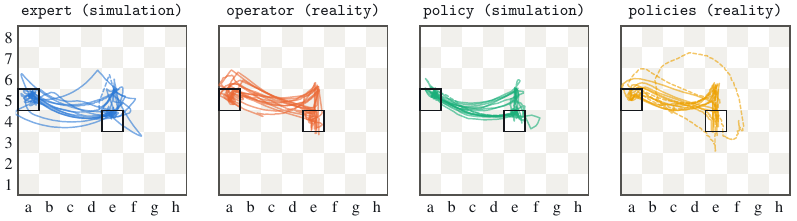}
  \caption{Gripper paths seen from above, from the randomized scripted expert, teleoperated demonstrations, trials of the simulation policy, and the first trial of eight of the real policies of Section~\ref{sec:real}. Dashed paths indicate unsuccessful trials.}
  \label{fig:paths}
\end{figure}

\label{sec:task}
The task is to pick up the king from one square of an otherwise empty chess board, place it on another square and return to the ready pose (Fig.~\ref{fig:stages}).
The source and target squares are given only in the instruction.
We study the task on a real robot arm and in its simulated counterpart.

\subsection{Real Arm}
\label{sec:robot}

\begin{figure}[t]
  \centering
  \includegraphics{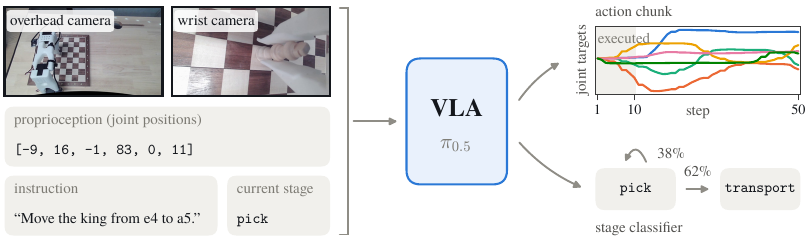}
  \caption{Inputs and outputs of the policy at one frame of a real trial. The policy observes the two rectified camera images, the joint positions, the instruction and the current stage. It outputs a chunk of 50 joint targets and predicts whether to transition to the next stage.}
  \label{fig:policy}
\end{figure}

The real setup consists of a low-cost 3D-printed arm with five joints and a gripper, the SO-101~\citep{so101,lerobot}, a second arm of the same type as the leader for teleoperation, and two cameras (Appendix~\ref{app:hardware}).
One camera is mounted on the wrist and looks along the gripper.
The other is fixed above the board on a 3D-printed overhead mount (Fig.~\ref{fig:policy}).
Both have wide-angle fisheye lenses, and the policy sees their images rectified to the simulator's pinhole model.
The board is wooden with 31\,mm squares.
The king is a wooden piece, 64\,mm tall on a 20\,mm base.

The task is difficult for this arm for four reasons.
First, the policy has to find each square from its name alone.
Second, the arm's posture varies considerably across the board, from steeply folded near the base to almost fully extended at the far ranks.
Third, the gripper has one fixed finger and one moving jaw, and a grasp that is a few millimeters off the king's axis topples the piece.
Fourth, the printed links flex under load by up to 2\,cm, so the arm cannot be positioned precisely without visual feedback.

\subsection{Simulator and Scripted Expert}
\label{sec:sim}

The simulator is a MuJoCo~\citep{mujoco} twin of the real setup.
It reproduces the arm's kinematics and servo dynamics, the measured pose of the board, the scanned king, and the poses and fields of view of the two cameras.

A motion planner would execute every move identically, which neither resembles teleoperation nor gives enough variation for imitation learning.
Our scripted expert therefore imitates the variability of a human operator.
It reaches for a target with a perception error, observes the remaining error after a latency and corrects, and repeats this until the error is below a tolerance.
Each episode samples its own tempo, errors, tolerances and grasp parameters, so paths and speeds differ between episodes (Fig.~\ref{fig:paths}).
Grasp errors arise from the same mechanism, and the expert detects a missed grasp, reopens and retries.
It succeeds in 93\% of its episodes.

The paths of the human operator on the real arm show a similar spread, although the operator's carries are straighter.
The fine-tuned policies vary less than both the expert and the operator, in simulation and on the real arm.

\subsection{Policy}
\label{sec:policy}

We start from the \pizerofive{} model, which combines a vision-language model with a flow-matching \emph{action expert} that attends to it.
The model was not pretrained on the SO-101 arm.
The policy receives the two camera views, the six joint positions, the instruction and the current stage.
It outputs chunks of 50 joint targets at 15\,Hz and executes 10 of them before planning again (Fig.~\ref{fig:policy}).
We fine-tune the action expert and the language model, 2.7 of the model's 4.1 billion parameters, and keep the vision encoder frozen.
The language model also learns to predict the current stage as one token (Fig.~\ref{fig:stages}), and the action expert attends to that token (Appendix~\ref{app:stage}).
At test time the policy re-scores the current and the next stage once per action chunk and advances when the next one wins twice in a row.

\section{Experiments}
\label{sec:experiments}

Our experiments address three questions, namely how behavior cloning scales with data and training (Section~\ref{sec:bc}), whether the policy can improve from its own rollouts (Section~\ref{sec:experience}), and how much simulation pretraining helps on the real arm (Section~\ref{sec:real}).
Appendix~\ref{app:tables} lists all simulation results.

\begin{figure}[t]
  \centering
  \includegraphics{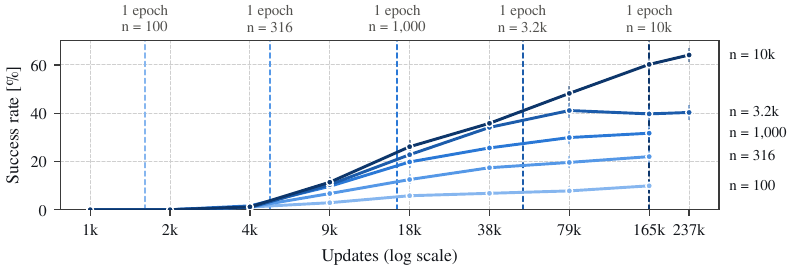}
  \caption{Success rate of the cooled-down models over updates, one curve per dataset size, with Wilson 95\% intervals. Dashed lines mark the end of each size's first epoch.}
  \label{fig:scaling}
\end{figure}

\subsection{Behavior Cloning in Simulation}
\label{sec:bc}

This section studies behavior cloning on the simulated demonstrations, where every model can be evaluated on many trials.

\textbf{Data.} The simulated demonstration set holds about 11.4k expert episodes with the piece offset by up to 8\,mm inside its square, of which 93\% are successes.
Training sets are nested subsets of the successes with $n = 100$, 316, 1,000, 3.2k and 10k episodes, and 30 further successes are held out for validation.

\textbf{Training.} Every run uses a constant learning rate of $3\cdot10^{-5}$ after 500 warm-up steps and a global batch of 32 frames.
The update budgets form a geometric ladder of eight steps from 1,000 to 165k (one epoch of the largest set), extended to 237k for the two largest sets.
For each dataset size, one constant-rate \emph{trunk} keeps a checkpoint at 90\% of every update budget, and a \emph{branch} from each checkpoint cools the rate to zero over the remaining updates~\citep{minicpm,hagele2024scaling}.
Each configuration is trained once.

\textbf{Evaluation.} Each model runs 1,000 closed-loop trials on a fixed schedule of moves, with the piece offset by up to 8\,mm as in the data, and with real-time chunking~\citep{rtc} at deployment settings.
A trial succeeds if the king ends standing with its center inside the target square and the arm is back at rest.
\emph{Strict success} also requires the king's whole base to be inside the square.
We report success rates with Wilson 95\% intervals, which are about $\pm3$ points at 30--60\%.

\textbf{Success increases with data and with updates.}\label{sec:scaling} Below about 4k updates no model succeeds, independent of the dataset size (Fig.~\ref{fig:scaling}).
Above that, success increases along both axes.
With 10k demonstrations it is still increasing at the largest budgets, from 60\% at 165k updates to 64\% at 237k.
At small update budgets, additional data has little effect (10\% and 11\% with 1,000 and 10k demonstrations at 9k updates).
In our grid, a smaller dataset performs as well as a larger one until its first epoch is complete, and performs worse afterwards.
Training for many epochs still helps, and in no case does it lower success beyond the evaluation noise.
In a partial ablation at 10k demonstrations, a model without \pizerofive{}'s robot pretraining reaches the same success rate (Appendix~\ref{app:pretrain}).\label{sec:pretrain}

\begin{figure}[t]
  \centering
  \includegraphics{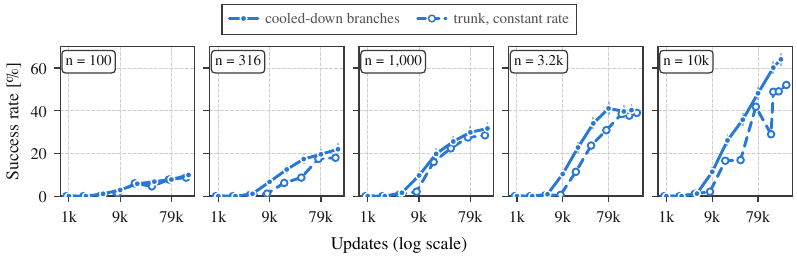}
  \caption{Constant-rate checkpoints against the cooled-down branches started from them.}
  \label{fig:cooldown}
\end{figure}

\textbf{A learning-rate cool-down is necessary for comparing checkpoints.}\label{sec:cooldown} A branch continues its trunk checkpoint for the last 10\% of the budget while the learning rate falls to zero (Fig.~\ref{fig:cooldown}).
With 10k demonstrations, the branch ending at 165k updates reaches 60\%, the constant-rate checkpoint it started from reaches 29\%, and the trunk's own checkpoint at 165k updates reaches 49\%.
The trunk's success rate is not monotonic in the number of updates, and early stopping on its checkpoints would select a model 18 points below what the run reaches with a cool-down.
Initial results suggest that SWA over the same updates is nearly as effective, with 57\% against 60\% for the cool-down at 10k demonstrations (Appendix~\ref{app:tables}).

\begin{figure}[t]
  \centering
  \includegraphics{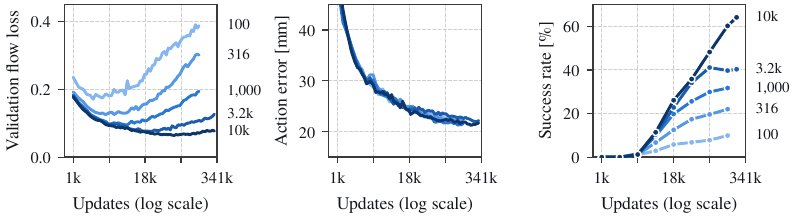}
  \caption{Validation loss (left) and open-loop action error (middle) of the constant-rate trunks, against the success rate of the cooled-down models (right). One curve per dataset size.}
  \label{fig:offline}
\end{figure}

\textbf{Offline metrics are unreliable for model selection.}\label{sec:offline} With 100 demonstrations the validation loss is lowest at about 2k updates and more than doubles by the end of training, while success increases from zero to 10\% (Fig.~\ref{fig:offline}).
Selection by validation loss would therefore return a model with zero success.
The open-loop action error does fall monotonically, but it does not discriminate between models.
At about 79k updates the error is 22 to 24\,mm for all five dataset sizes, while the success rate of the cooled-down models ranges from 8\% to 48\%.

\begin{figure}[t]
  \centering
  \includegraphics{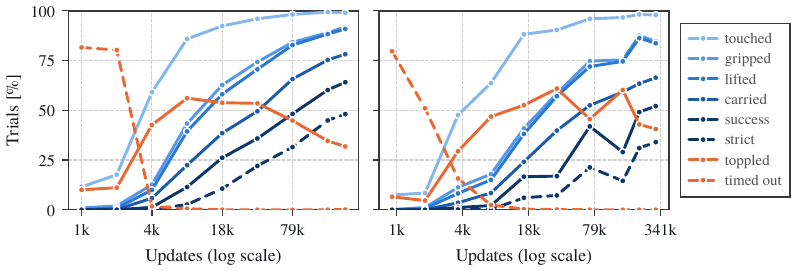}
  \caption{Share of the 1,000 trials that reach each milestone (blue, strict success dashed) and that topple the piece or time out (orange, timeouts dashed), at 10k demonstrations. Left: cooled-down models. Right: constant-rate checkpoints.}
  \label{fig:funnel}
\end{figure}

\textbf{Where trials fail.} The milestones are acquired in order during training (Fig.~\ref{fig:funnel}).
Touching the piece comes first, gripping and lifting follow together, and placing comes last and is still improving at the largest budget.
Most failures after a grip are toppled pieces, in about a third of all trials.
The cool-down mostly improves the grasp and the placement.

\subsection{Learning from the Policy's Own Rollouts}
\label{sec:experience}

In this section, we evaluate whether the policy improves when it is trained on its own rollouts in addition to the demonstrations.

\textbf{Setup.} We take the best behavior-cloning model of a dataset size (165k updates), run it on new moves from the collection distribution and record every trial with its graded outcome.
A new policy is trained from the base model on the demonstrations plus the rollouts, on the same update ladder.
There are two rounds.
1K+1K uses 1,000 demonstrations and 1,000 rollouts of the 1,000-demonstration model (32\% success).
10K+10K uses 10k demonstrations and 10k rollouts of the 10k-demonstration model (60\% success).
\begin{figure}[t]
  \centering
  \includegraphics{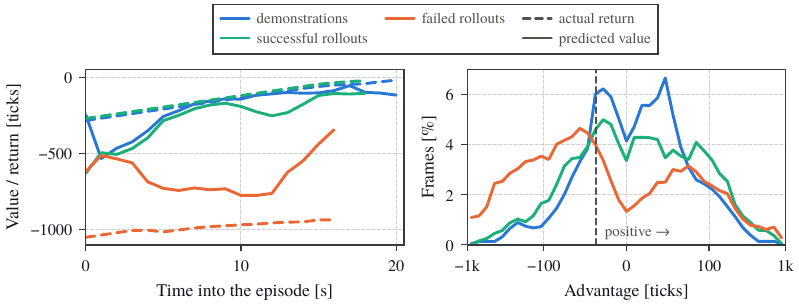}
  \caption{The value function of the 1K+1K round. Left: predicted value (solid) and actual return (dashed) on held-out episodes. Right: distribution of the advantage over the training frames, on a symmetric-log axis. Rollout frames right of the dashed threshold are labeled positive; demonstration frames are labeled positive regardless. One tick is one control step, 1/15\,s.}
  \label{fig:value}
\end{figure}

We compare four variants.
\emph{Filtered BC} trains on the demonstrations plus the successful rollouts.
RECAP-style advantage conditioning~\citep{pistar06} labels every frame with a learned value function.
A cross-fitted variant of it labels each half of the episodes with a value function that never saw it.
\emph{Outcome labels} mark every frame of a success as positive and every frame of a failure as negative.
We sample advantage-conditioned policies with the positive label, optionally with classifier-free guidance~\citep{cfg}.
The baseline is the model trained on the demonstrations alone for the same number of updates.
A large value function trained for several epochs memorized its training episodes in our setting, which makes its labels uninformative.
We therefore use a smaller value function, train it for one epoch and monitor it on held-out rollouts (Fig.~\ref{fig:value}).
Appendix~\ref{app:experience} gives the details and all numbers.

\begin{figure}[t]
  \centering
  \includegraphics{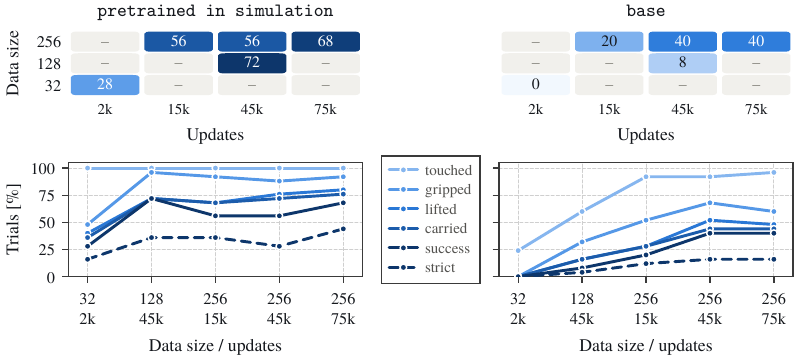}
  \caption{Partial results in the real setting, 25 trials per model. Left: policies initialized from the simulation policy. Right: policies initialized from the base \pizerofive{} model. Top: success rate (\%) over demonstrations and updates, where a dash marks a model without a full evaluation. Bottom: share of trials that reach each milestone, per evaluated model.}
  \label{fig:real}
\end{figure}

\textbf{Results.} No variant improves over behavior cloning beyond the evaluation noise (Appendix~\ref{app:experience}, Table~\ref{tab:all}).
The three advantage-conditioned variants match behavior cloning.
The policy does use the label, since success drops from 30\% with the positive label to 24\% with the negative label.
Without a label it reaches 26\%, guidance at $\beta = 1.5$ gives 31\%, and guidance at $\beta = 2.5$ lowers success to 23\% (Appendix~\ref{app:experience}, Fig.~\ref{fig:guidance}).
Filtered behavior cloning is about three points ahead of behavior cloning at two intermediate update budgets, but not at the largest one.
The failed rollouts, which only the labeled variants see, provide no measurable benefit.

\textbf{Corrections and residual reinforcement learning.} In the original RECAP, a person corrects the policy during its rollouts.
We reproduce this in simulation with an automatic coach.
A monitor on the simulator's state hands control to the scripted expert when a failure is imminent, and the expert re-grasps or places the king and returns control to the policy.
Training on about 1,000 coached rollouts gives no consistent gain.
The best coached variant is above behavior cloning on one evaluation schedule and below it on the other.
Residual reinforcement learning, in which a small network trained with PPO adds joint offsets to the frozen policy's actions, did not improve success in any of four runs.

\subsection{Experiments on the Real Arm}
\label{sec:real}

On the real arm of Section~\ref{sec:robot}, we measure how much real data the policy needs and whether pretraining in simulation helps.

\textbf{Data.} One of the authors teleoperated 261 king moves with the leader arm, each starting and ending at the ready pose.
The source and target squares were drawn at random, and the king was placed up to 8\,mm off the center of its square, as in the simulated data.
259 of the moves are successes, and these form the real dataset, about half an hour of motion in total.
The stage labels are detected from the recorded trace with the simulator's own definitions.

\textbf{Training.} We fine-tune \pizerofive{} on nested random subsets of 32, 128 and 256 demonstrations, from two initializations.
\emph{Base} is the released \pizerofive{} model.
\emph{Pretrained in simulation} is our \pizerofive{} simulation model trained on 10k demonstrations for 237k updates and cooled down (64\% success in simulation).
We call the policies fine-tuned from it sim-initialized and those fine-tuned from the base \pizerofive{} model base-initialized.
The recipe is the same as in simulation (Section~\ref{sec:bc}).
A constant-rate trunk runs to 45k updates and keeps checkpoints at 90\% of the budgets 2k, 15k and 45k updates, and a branch cools down from each checkpoint.
The trunks are also extended to 75k updates, with one more cooled-down model per trunk.

\textbf{Evaluation.} Every model runs the same 25 trials, the first 25 moves of the simulator's easy schedule, with the king centered on its square.
The policy runs each trial autonomously, at one third of the demonstrated tempo because of the inference latency in our deployment.
A person resets the king and grades the highest milestone reached, with the same definitions of success and strict success as in simulation (Appendix~\ref{app:realeval}).
With 25 trials per model, we only interpret large differences.

\textbf{Results.} The results are partial, with ten models evaluated in full (Fig.~\ref{fig:real}, Appendix~\ref{app:tables}, Table~\ref{tab:real}).
In all five evaluated pairs the sim-initialized policy is better, and with little real data the difference is large.
With 128 demonstrations, the sim-initialized policy succeeds in 18 of 25 trials and the base-initialized policy succeeds in 2.
With 32 demonstrations, the sim-initialized policy succeeds in 7 trials, while the base-initialized policy never grips the king.
With 256 demonstrations the difference is smaller.
Depending on the update budget, the sim-initialized policy succeeds in 14 to 17 trials and the base-initialized policy in 5 to 10 trials.

\section{Related Work}
\label{sec:related}

\textbf{VLAs and their adaptation.} $\pi_0$~\citep{pi0} introduced the pairing of a vision-language model with a flow-matching action expert, and \pizerofive{}~\citep{pi05} extended it with co-training on heterogeneous data for open-world generalization.
SmolVLA~\citep{smolvla} showed that a small open model trained on community data is competitive.
\citet{lin2024datascaling} measured data scaling in imitation learning for policies trained from scratch.

\textbf{Simulation as pretraining.} Simulated demonstrations are a cheap complement to real ones.
Co-training a policy on both, with the simulator roughly matched to the real scene, raises real success well beyond what the real data alone gives~\citep{cotrain}, and large simulated corpora are part of the pretraining mix of recent generalist models~\citep{gr00t}.

\textbf{Learning-rate schedules.} Warmup-stable-decay schedules~\citep{minicpm} train at a constant rate and cool down briefly at the end.
\citet{hagele2024scaling} showed that cool-downs branched off one constant-rate run match full cosine runs at every length, so one run per dataset size is enough for a whole update axis.
They also suggested SWA~\citep{swa} as an alternative to the cool-down.

\textbf{Learning from experience.} RECAP~\citep{pistar06} trains a value function on the policy's own rollouts and corrections, labels each frame by its advantage, conditions the policy on a text indicator of that label, and steers with classifier-free guidance~\citep{cfg} at test time.
It improved throughput and robustness on long real-world tasks.
Filtered behavior cloning, which keeps only the successful rollouts, is a simple baseline for learning from rollouts.

\section{Conclusion}
\label{sec:discussion}

We studied the fine-tuning of a VLA for a chess-move task with real and simulated demonstrations.
In our setting, fine-tuning choices could only be judged by closed-loop evaluation on many trials, which the simulator made affordable.
The same simulator also provided useful pretraining data for the real arm.

\textbf{Limitations.} Our results come from one task and with a single training run per setting.
The scripted expert is randomized but its similarity to human demonstrations is not rigorously measured.
Our RECAP variants omit the original's advantage-conditioned pretraining and human corrections.
The statistical evidence from the real arm is limited, with 25 trials per model.

\clearpage
\acknowledgments{
This project was supported by the Charles University Grant Agency (GAUK), project no.~458326.
This research was partially supported by SVV project number 260\,821.
We acknowledge VSB -- Technical University of Ostrava, IT4Innovations National Supercomputing Center, Czech Republic, for awarding this project access to the LUMI supercomputer, owned by the EuroHPC Joint Undertaking, hosted by CSC (Finland) and the LUMI consortium through the Ministry of Education, Youth and Sports of the Czech Republic through the e-INFRA CZ (grant ID: 90254).
}

\bibliography{refs}

\appendix
\raggedbottom
\section{Setup of the Real Arm}
\label{app:hardware}

\textbf{Hardware.} The arm is the SO-101 of The Robot Studio and LeRobot~\citep{so101,lerobot}, with a follower and a leader arm.
The gripper consists of a fixed finger and a moving jaw with flat pads.
The two cameras are identical USB modules with a 130$^\circ$ diagonal field of view, recorded at $1280 \times 720$.
The follower and both cameras connect to a laptop that runs the control loop, and the policy is served from a remote GPU.
The simulator uses the arm's URDF and servo dynamics, the measured board pose, the king's scanned profile, and the calibrated camera poses.
It does not model the compliance of the printed links.

\begin{figure}[H]
  \centering
  \includegraphics{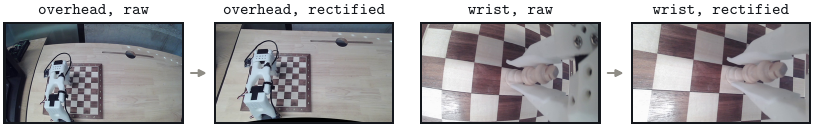}
  \caption{Both camera frames as recorded (raw) and rectified to the simulator's pinhole model, as the policy sees them.}
  \label{fig:rectify}
\end{figure}

\textbf{Sim-to-real differences.} The real cameras have fisheye lenses and the simulator's cameras are pinhole cameras.
We record the real frames raw and rectify them to the simulator's intrinsics in training and deployment (Fig.~\ref{fig:rectify}).
The joint zeros of the real calibration differ from the simulator's by a constant offset per joint, which normalization with the real dataset's statistics absorbs.
Two differences are not handled.
The real policy runs at a third of the demonstrated tempo, and the simulator applies the 15\,Hz joint targets as steps where the real controller interpolates them.

\textbf{Collection.} The operator sees the next move on a screen, places the king, and moves it with the leader arm, starting and ending at the ready pose.
The recorded frame rate varied between sessions (15 to 23 frames per second, stored as 30), so the policies were trained on a mix of apparent tempos.

\section{Result Tables}
\label{app:tables}

This appendix lists the numbers behind the figures of Section~\ref{sec:experiments}.

\begin{table}[H]
  \centering
  \small
  \caption{Simulation results: success rate (\% of 1,000 trials). Top: behavior cloning, with models cooled down unless noted. Bottom: learning from 1,000 of the policy's own rollouts.}
  \label{tab:all}
  \begin{tabular}{@{}lrrr@{}}
    \toprule
    Model & Demos & Updates & Success \\
    \midrule
    \pizerofive{}, behavior cloning & 100 & 165k & 9.9 \\
    \pizerofive{}, behavior cloning & 316 & 165k & 22.0 \\
    \pizerofive{}, behavior cloning & 1,000 & 165k & 31.7 \\
    \pizerofive{}, behavior cloning & 3.2k & 165k & 39.7 \\
    \pizerofive{}, behavior cloning & 10k & 165k & 60.2 \\
    \pizerofive{}, behavior cloning & 10k & 237k & 64.1 \\
    \quad constant rate, no cool-down & 10k & 165k & 48.9 \\
    \quad SWA, no cool-down & 10k & 165k & 57.0 \\
    \quad no robot pretraining & 10k & 165k & 59.5 \\
    $\pi_0$, stock recipe & 10k & 165k & 77.7 \\
    SmolVLA, stock recipe & 10k & 165k & 54.5 \\
    \midrule
    Behavior cloning, demonstrations only & 1,000 & 79k & 29.9 \\
    Filtered BC (+ successful rollouts) & 1,000 & 79k & 32.9 \\
    RECAP, value function & 1,000 & 79k & 30.0 \\
    RECAP, cross-fitted value function & 1,000 & 79k & 31.3 \\
    RECAP, outcome labels & 1,000 & 79k & 29.8 \\
    RECAP, outcome labels, guidance $\beta = 1.5$ & 1,000 & 79k & 31.4 \\
    RECAP, coached rollouts & 1,000 & 79k & 34.7 \\
    Filtered BC, coached rollouts & 1,000 & 79k & 28.9 \\
    \bottomrule
  \end{tabular}
\end{table}

\begin{table}[H]
  \centering
  \small
  \caption{Evaluation on the real arm, as the number of the 25 trials that reached each milestone.}
  \label{tab:real}
  \setlength{\tabcolsep}{5pt}
  \begin{tabular}{@{}rr@{\hspace{14pt}}rrr@{\hspace{14pt}}rrr@{}}
    \toprule
    & & \multicolumn{3}{c}{Pretrained in simulation} & \multicolumn{3}{c}{Base} \\
    \cmidrule(r{10pt}){3-5} \cmidrule{6-8}
    Demos & Updates & Touched & Gripped & Success & Touched & Gripped & Success \\
    \midrule
    32 & 2k & 25 & 12 & 7 & 6 & 0 & 0 \\
    128 & 45k & 25 & 24 & 18 & 15 & 8 & 2 \\
    256 & 15k & 25 & 23 & 14 & 23 & 13 & 5 \\
    256 & 45k & 25 & 22 & 14 & 23 & 17 & 10 \\
    256 & 75k & 25 & 23 & 17 & 24 & 15 & 10 \\
    \bottomrule
  \end{tabular}
\end{table}

\section{Stage Prediction}
\label{app:stage}

\begin{figure}[H]
  \centering
  \includegraphics{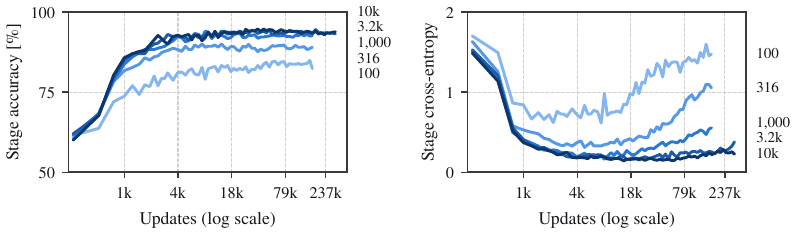}
  \caption{Accuracy (left) and cross-entropy (right) of the predicted stage on held-out demonstrations.}
  \label{fig:stage}
\end{figure}

The stage accuracy is close to its final value after about 4k updates for every dataset size, long before the success rate rises (Fig.~\ref{fig:stage}).

\section{Value Function Training}
\label{app:value}

\begin{figure}[H]
  \centering
  \includegraphics{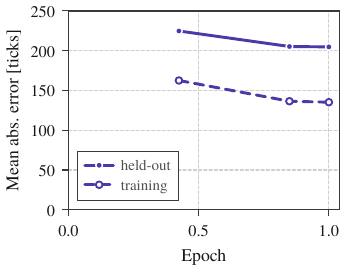}
  \caption{Error of the 1K+1K value function during its single training epoch, on held-out and on training episodes. One tick is 1/15\,s.}
  \label{fig:valuetrain}
\end{figure}

At the end of the epoch, the error is 205 ticks on held-out episodes and 135 ticks on training episodes (Fig.~\ref{fig:valuetrain}).
Failed episodes account for most of it, with 375 ticks held-out against 106 ticks on successful episodes.

\section{Details on Learning from Rollouts}
\label{app:experience}

\begin{figure}[H]
  \centering
  \includegraphics{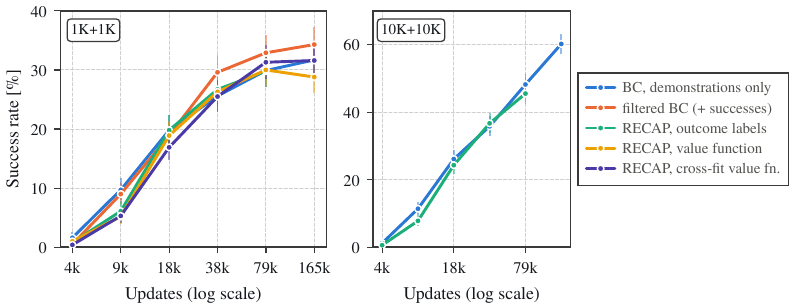}
  \caption{Learning from the policy's own rollouts, cooled-down models, each sampled with the positive label.}
  \label{fig:experience}
\end{figure}

Figure~\ref{fig:experience} shows the variants of Section~\ref{sec:experience} at every update budget they were trained for.

\begin{figure}[H]
  \centering
  \includegraphics{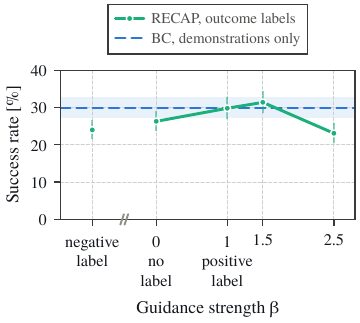}
  \caption{Success of the outcome-label RECAP policy at 79k updates, sampled with different labels and guidance strengths $\beta$. Shaded: the Wilson 95\% interval of behavior cloning.}
  \label{fig:guidance}
\end{figure}

\textbf{Variants.} All variants train on the demonstrations plus the policy's rollouts and are compared with behavior cloning on the demonstrations alone at the same number of updates.
\begin{itemize}
  \item \emph{Filtered BC} adds only the successful rollouts.
  \item \emph{RECAP, value function} follows RECAP~\citep{pistar06} with a 500M value function trained for one epoch. A frame's advantage is the progress made over the next second minus the progress the value function expects. The threshold is the 30th percentile of the advantages of all training frames, demonstrations included. Rollout frames above it are labeled positive, and every demonstration frame is labeled positive. This labels 72\% of the frames of successful rollouts and 48\% of those of failed rollouts positive.
  \item \emph{RECAP, cross-fitted} labels each half of the episodes with a value function that never saw it.
  \item \emph{RECAP, outcome labels} marks every frame of a success as positive and every frame of a failure as negative.
\end{itemize}
Advantage-conditioned policies are sampled with the positive label, and with classifier-free guidance~\citep{cfg} at $\beta \in \{1.5, 2.5\}$.

\textbf{Value function.} A value function with the policy's 3B backbone, trained for two epochs, memorized its training episodes and was at chance on unseen ones.
The smaller value function used here is trained for one epoch and monitored on held-out rollouts (Fig.~\ref{fig:value}, Appendix~\ref{app:value}).
On held-out episodes it separates successes from failures with an AUC of 0.85 four seconds into the episode, which is before the grasp.

\textbf{Results.} At 79k updates in the 1K+1K round, behavior cloning reaches 30\%, the three advantage-conditioned variants reach 30 to 31\%, and filtered BC reaches 33\% (Table~\ref{tab:all}).
At 165k updates, filtered BC reaches 34\% against 32\% for behavior cloning, and the advantage-conditioned variants reach 29 to 32\%.
Sampled with the negative label, without a label and with the positive label, the outcome-label model reaches 24\%, 26\% and 30\% (Fig.~\ref{fig:guidance}).
In the 10K+10K round, outcome labels reach 46\% at 79k updates against 48\% for behavior cloning.

\textbf{Corrections from an automatic coach.} A monitor on the simulator's state hands control to the scripted expert when the king tilts or slides, when a grasp misses, or when a forward simulation of the queued actions predicts a topple.
We train on about 1,000 coached rollouts with RECAP labels and as filtered BC on the coached successes.
At 79k updates coached RECAP reaches 35\% on the hard schedule, above behavior cloning (30\%, paired McNemar test, $p < 0.02$), but on the easy schedule it is below behavior cloning (46\% against 50\%).

\textbf{Residual reinforcement learning.} A small network trained with PPO from privileged simulator state adds a joint offset of at most $3^\circ$ to the frozen policy's actions.
In four runs with different rewards and exploration noise, the success rate stayed within evaluation noise of its initial value.

\section{Pretraining Ablation}
\label{app:pretrain}

\begin{figure}[H]
  \centering
  \includegraphics{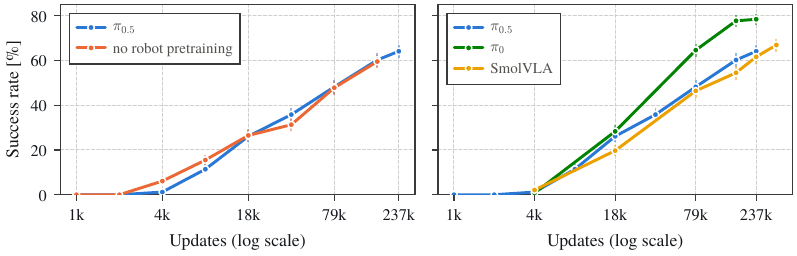}
  \caption{Partial results of the pretraining ablation, for cooled-down models at 10k demonstrations. Left: \pizerofive{} with and without robot pretraining, trained with the same recipe. Right: other model families with their stock recipes.}
  \label{fig:pretrain}
\end{figure}

We initialize the vision-language model from PaliGemma, on which \pizerofive{} is built, and initialize the action expert and the projections randomly.
Trained with the same recipe on 10k demonstrations, this model follows the same curve as \pizerofive{} and also reaches 60\% at 165k updates (Fig.~\ref{fig:pretrain}).
This result is partial, since the ablation covers only the largest dataset size, and robot pretraining may matter with smaller datasets.
Other model families, fine-tuned with their stock recipes, perform in the same range or better.
At 165k updates, $\pi_0$~\citep{pi0} reaches 78\% and SmolVLA~\citep{smolvla} reaches 55\%.
Their recipes differ from ours, so these numbers compare model and recipe together.

\section{Released Dataset and Environment}
\label{app:data}

The real demonstrations are in the LeRobot format (Table~\ref{tab:data}).
The simulated demonstrations can be regenerated with the released environment and scripted expert.\\
The code is at \url{https://github.com/kripner/vla-chess} and the dataset at\\ \url{https://huggingface.co/datasets/Kripi/so101-chess-king-moves-real}.

\begin{table}[H]
  \centering
  \small
  \caption{Released dataset.}
  \label{tab:data}
  \begin{tabular}{@{}llrl@{}}
    \toprule
    Dataset & Source & Episodes & Contents \\
    \midrule
    king moves, real & teleoperation & 261 & the demonstrations of Section~\ref{sec:real} \\
    \bottomrule
  \end{tabular}
\end{table}

\section{Evaluation Protocol on the Real Arm}
\label{app:realeval}

Every model is evaluated on the same 25 trials, the first 25 moves of the simulated \emph{easy} schedule, in which each trial's target square is the next trial's source and the king starts centered on its square.
A trial starts with the arm at its ready pose and ends when the policy switches to its final stage or after 60\,s of control time, which counts as a failure.
The human evaluator only resets the king and grades the highest milestone reached (Table~\ref{tab:realfunnel}).
A failed trial is never discarded, and a trial is repeated only for a setup fault, such as the king placed on the wrong square.
Every trial is recorded, whether it succeeds or not.

\begin{table}[H]
  \centering
  \small
  \caption{Milestones graded by the evaluator. Success is milestone~5 or~6, strict success milestone~6.}
  \label{tab:realfunnel}
  \begin{tabular}{@{}rll@{}}
    \toprule
    & Milestone & Observed \\
    \midrule
    0 & none & the jaws never touched the king \\
    1 & touched & a jaw touched the king \\
    2 & gripped & the king was between both jaws, touching both \\
    3 & lifted & gripped and lifted clear of the board (about 1.5\,cm) \\
    4 & over target & carried until its center was above the target square \\
    5 & success & at the end the king stands, released, its center inside the target square \\
    6 & strict success & as 5, and its base lies entirely inside the square \\
    \bottomrule
  \end{tabular}
\end{table}

\end{document}